\documentclass[letterpaper, 10 pt, conference]{ieeeconf}  % Comment this line out if you need a4paper

\IEEEoverridecommandlockouts                              % This command is only needed if 
\usepackage{graphics} % for pdf, bitmapped graphics files
\usepackage{epsfig} % for postscript graphics files
\usepackage{mathptmx} % assumes new font selection scheme installed
\usepackage{times} % assumes new font selection scheme installed
\usepackage{amsmath} % assumes amsmath package installed
\usepackage{amssymb}  % assumes amsmath package installed
\usepackage{verbatim} 
\usepackage{textcomp}
\usepackage{url}
\usepackage{booktabs,chemformula} % For nicer tables

\usepackage{gensymb}
\usepackage{titlesec} % for section formatting and counter handling
\usepackage{multirow}

\usepackage{hyperref}

\usepackage{caption}
\usepackage{subcaption}
\usepackage{siunitx}
\usepackage{tikz}
\usepackage{wasysym}
\usepackage{svg}

\usepackage{ulem}
\usepackage{soul}
\usepackage{xcolor}

\usepackage{pifont}% http://ctan.org/pkg/pifont

\usepackage[font=small]{caption}
\usepackage[outline]{contour}

\usepackage{lipsum}
\usepackage{hyperref}

\usepackage{tikz}

\usepackage[]{svg}

\usepackage{multirow} % For multirow support in tables
\usepackage{makecell} % For line breaks and alignment in table cells

\usepackage{array}

\usepackage{colortbl} % For table coloring

\usepackage{hhline}

\usepackage{diagbox}

\usepackage{hyperref} % Load hyperref first
\usepackage{adjustbox} % include in preamble

\usepackage{cite}
\usepackage{comment}

\usepackage[utf8]{inputenc}
\DeclareUnicodeCharacter{2605}{\starr}
\definecolor{applegreen}{rgb}{0.55, 0.71, 0.0}

\newcommand{\lilly}[1]{\textcolor{red}{LTC: #1}}

\newbox{\bigpicturebox}

\title{\LARGE \bf
BiFlex: A Passive Bimodal Stiffness Flexible Wrist \\ for Manipulation in Unstructured Environments
}

\author{Gu-Cheol Jeong$^{1}$, Stefano Dalla Gasperina$^{1}$, Ashish D. Deshpande$^{1,2}$, Lillian Chin$^{1,*}$ and Roberto Mart{\'i}n-Mart{\'i}n$^{1,*}$%
\thanks{$^1$The University of Texas, Austin, TX, USA}%
\thanks{$^2$Meta Reality Labs Research, Redmond, WA, USA.}%
\thanks{$^*$These authors provided equal advising}
}

\begin{document}

\maketitle
\thispagestyle{empty}
\pagestyle{empty}

%%%%%%%%%%%%%%%%%%%%%%%
\begin{abstract}
Robotic manipulation in unstructured, human-centric environments poses a dual challenge: achieving the precision need for delicate free-space operation while ensuring safety during unexpected contact events. Traditional wrists struggle to balance these demands, often relying on complex control schemes or complicated mechanical designs to mitigate potential damage from force overload.
In response, we present BiFlex, a flexible robotic wrist that uses a soft buckling honeycomb structure to provides a natural bimodal stiffness response. The higher stiffness mode enables  precise household object manipulation, while the lower stiffness mode provides the compliance needed to adapt to external forces.
We design BiFlex to maintain a fingertip deflection of less than 1 cm while supporting loads up to 500g and create a BiFlex wrist for many grippers, including Panda, Robotiq, and BaRiFlex. We validate BiFlex under several real-world experimental evaluations, including surface wiping, precise pick-and-place, and grasping under environmental constraints. We demonstrate that BiFlex simplifies control while maintaining precise object manipulation and enhanced safety in real-world applications. More information and videos at {\url{https://robin-lab.cs.utexas.edu/BiFlex/}}
\end{abstract}

% designing a flexible robot wrist that is specifically tailored for interaction in contact rich tasks within human environment. 
% Our innovative design addresses the challenges poses by unexpected force overload encountered during such tasks, offering features like bimodal stiffness, universal compatibility, and simplicity. 
% This advancement leverages the buckling effect, allowing the wrist to switch its stiffness in response to external disturbances. 
% Additionally, the use of soft materials provides inherent elasticity, accommodating deformations during contact-rich interaction.
% With our wrist, we achieve \lilly{results} 
%%%%%%%%%%%%%%%%%%%%%%%
\section{Introduction}
\label{s:intro}

% In unstructured environments (homes, offices, \ldots), simple physical tasks such as wiping a table demand that robotic agents create and control contact-rich interactions.
% While other tasks, such as pick-and-place, can be successfully performed just by controlling motion, contact-rich interactions require modulating the amount of force the robot applies.

Designing robots capable of physical tasks in unstructured environments remains one of the core open problems of modern robotics. 
Unstructured settings are characterized by their inherent uncertainty that exposes robotic end-effectors to frequent and unpredictable forces.
For example, when grasping a flat object or wiping a surface, inaccuracies in the perceived location could lead to the robot missing the target, or creating unexpected and dangerously high reactive forces that could damage the robot.

\begin{figure}
    \centering
    \begin{subfigure}[t]{\columnwidth}
        \centering        \includegraphics[width=\textwidth]{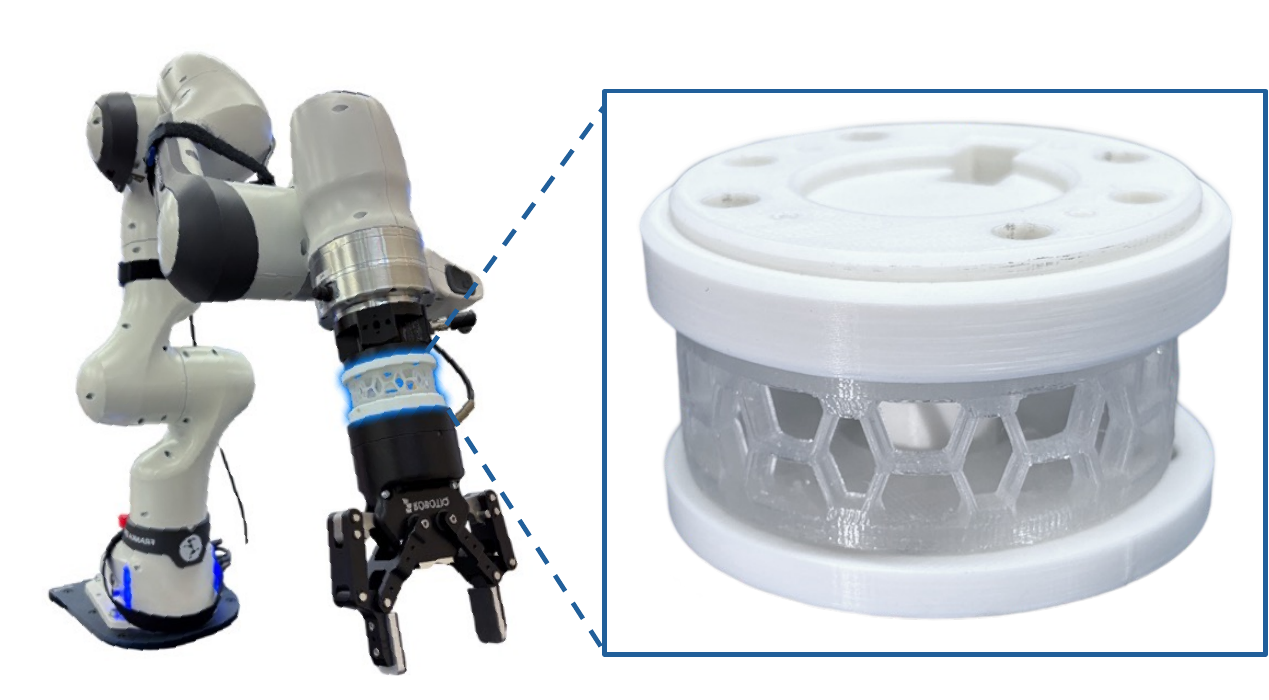}
    \end{subfigure}
    \begin{subfigure}[t]{\columnwidth}
        \centering
        \includegraphics[width=\textwidth]{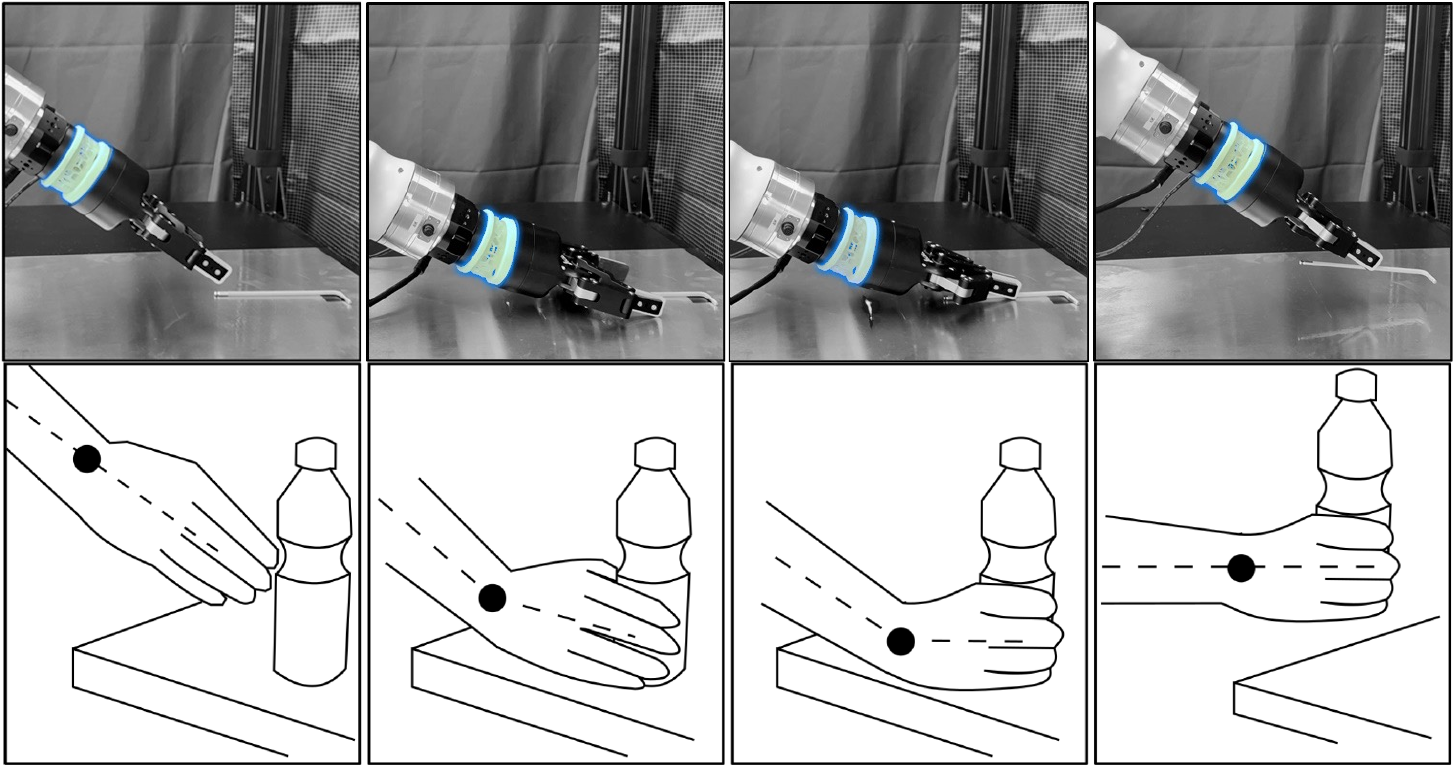}
    \end{subfigure}
    \caption{(top) The BiFlex wrist, featuring a honeycomb structure, mounted on a robotic arm and attached to a robotic gripper. (bottom) Typical manipulation sequence using the BiFlex wrist. When grasping in unstructured environments, the wrist transitions from high- to low-stiffness states upon contact with the table surface to exploit it as a constraint. Illustration adapted from~\cite{kanitz_compliant_2018}. }
    \vspace{-4mm}
   \label{fig:teaser}
   
\end{figure}

To overcome these high-precision demands, humans exploit the environment as an additional constraint~\cite{della2017postural,mason1985mechanics,erdmann1988exploration,eppner2015exploitation,kanitz_compliant_2018}. By flexing or tensing our wrists, we dynamically change our morphology to adapt to different environmental conditions (Fig.~\ref{fig:teaser}). For example, to grasp a water bottle from a table, we keep a stiff wrist to ensure precise object grasping and then flex our wrist when we come into contact with the table. Building a similarly robust strategy for traditional rigid robot arms would require computationally complex force control / feedback architectures~\cite{siciliano2008springer,siciliano2009force,khatib1987unified,hogan1984impedance}.

% This simplifies manipulation through a dynamic behavior that adapts their morphology to high reactive forces (flexing when contacting the table) while resisting lower forces (the weight of the object) to enable precise subsequent object placement (Fig.~\ref{fig1:teaser}, bottom). 
% through dynamic contact-rich interactions. 
% A critical challenge in these settings is the variable resistance to forces: while some tasks, such as precise object placing, require to  
% in these conditions requires both accurate positioning and the ability to handle unexpected external forces. 

A promising alternative to this complex computation is outsourcing the morphology change to the embodiment itself, especially by varying compliance at the wrist~\cite{montagnani2015finger,phan2020estimating, bajaj_state_2019,fan_prosthetic_2022}. 
% In human manipulation, the wrist can shift from a high stiffness state for accurate precise placement to a low stiffness state for safe interactions with the environment (Fig.~\ref{fig1:teaser}B). 
% Inspired by this approach, robotics researchers have sought to vary compliance, either by having rigid wrists react to external forces through control algorithms~\cite{calanca_review_2016}, 
This can be achieved by developing wrist mechanisms that can actively change their stiffness~\cite{kanitz_compliant_2018,zhang_stiffness_2020,von2020compact} or by directly incorporating compliant elements directly into the wrist's mechanical design~\cite{zhang_0_2024}. Both approaches have drawbacks. Active stiffness mechanisms add complexity at the mechanical and algorithmic level, increasing bulk andcost. Meanwhile, compliant wrists sacrifice precision in pick-and-place operations in order to better react to external forces. Both approaches require a custom installation for a specific end effector, limiting the wrist's generalizability across available robot arms and hands.

We address these shortcomings through BiFlex, a passive robotic wrist (Fig.~\ref{fig:teaser}) that can seamlessly shift between rigid and compliant modes without adding unnecessary complexity to the system. BiFlex leverages the buckling effect to shift between two distinct stiffness modes. This ``bimodal'' stiffness provides both rigid and compliant behavior without requiring additional actuators or sensors (Fig.~\ref{fig2:desired_behavior}). To create BiFlex, we design a compact and lightweight 3D-printed honeycomb structure, ensuring safe and stable operation during interaction, while maintaining accuracy during free-space movements. BiFlex's simple design means it can be easily customized to fit a specific end-effector's performance requirements. 

In this paper, we contribute a theoretical analysis of BiFlex's bimodal stiffness behavior and validate the honeycomb structure's buckling effect. We use this analysis to create BiFlex wrists for 3 different robot hands, demonstrating the versatility of our design.We use the BiFlex wrist to successfully perform several contact-rich tasks, such as wiping an unknown non-flat surface and grasping flat objects, while maintaining precise free-space motion. We consider BiFlex a step towards a new type of robot embodiment that can facilitate safe and computationally simpler manipulation in uncertain, unstructured environments.

%Experimental results show that our wrist design enables a robotic manipulator to successfully perform contact-rich tasks, such as wiping non-flat surfaces even with uncertainty about the surface height, while maintaining precise free-space motion and grasping flat objects despite position uncertainty. BiFlex with the gripper was capable of wiping a triangular hill surface up to \SI{50}{mm} in height while maintaining an applied force below \SI{15}{N}, and it successfully handled 14 out of 15 household objects in our pick-and-place experiments.

%The BiFlex offers precision, adaptability, and low weight in a simple, passive design for any robot arm.

%Finally, we experimentally demonstrate the performance of the proposed wrist in contact-rich environments, showing that a robotic manipulator can successfully wipe a non-flat surface while guaranteeing accurate positioning in free-space motion, and grasp a flat object even with uncertainty in its position.

%%%%%%%%%%%%%%%%%%%%%%%
\section{Related Work}
\label{s:related_work}

\input{Figure_tex/fig2_desired_behavior}

To address the limitations of rigid / fixed stiffness wrists, researchers have investigated variable stiffness mechanisms to modulate wrist compliance. These designs mimic the adaptability of human biomechanics, where humans control wrist stiffness through muscular co-contraction~\cite{formica2012passive,borzelli2018muscle}. Variable stiffness mechanisms adapt to specific task requirements, providing higher stiffness for precision tasks and lower stiffness for tasks that require exploration or compliance~\cite{sun_compact_2025}. 

Fully actuated wrists are a common approach for modulating stiffness~\cite{milazzo2024modeling,sun_compact_2025,bajaj_state_2019}. These wrists use electric or pneumatic actuators to actively and continuously control the stiffness at any given time. For example, Milazzo et al.~\cite{milazzo2024modeling} uses four electric motors to control wrist stiffness over 3 degrees-of-freedom (DoF), while Sun et al.~\cite{sun_compact_2025} uses tendons to drive a 2 DoF wrist. These wrists require significant numbers of actuators and electronics to achieve their active control, limiting their application beyond the laboratory setting.

Underactuated bimodal stiffness wrists have been proposed as a simplified alternative to fully actuated wrists~\cite{montagnani_preliminary_2013,von2020compact,zhang_stiffness_2020}. These systems leverage mechanical principles to switch between two discrete stiffness states: typically, a low-stiffness mode for adaptability and a high-stiffness mode for precision ~\cite{montagnani_preliminary_2013}. For example, 
Von Drigalski et al.~\cite{von2020compact} introduced a compact, cable-driven soft wrist with a pneumatically-driven locking mechanism to switch between rigid and compliant states, while Zhang et al.~\cite{zhang_stiffness_2020} uses pneumatics with a reconfigurable elastic inner skeleton and an origami shell. These solutions still require power sources, active switching  mechanisms, and increased weight.

%Both fully actuated and underactuated wrist designs offer solutions for variable stiffness that provide different levels of adaptability, but they come with significant challenges. Fully actuated systems provide continuous stiffness modulation through multiple actuators, yet they are often complex, heavy and require an external power source. While underactuated bimodal stiffness solutions are simpler and sufficient for many applications, they still  Both approaches balance adaptability and precision, but integrating them into existing robotic systems remains difficult. 

In this paper, we address these limitations through a fully passive wrist design that achieves bimodal stiffness without actuators. This design offers simplicity, reduced cost, and compactness, while eliminating the complexity and weight of active systems.

%%%%%%%%%%%%%%%%%%%%%%%
\section{BiFlex Wrist Design}
\label{s:method}

We aim to design a wrist that transitions between a rigid state for precision tasks and a compliant state for safe interaction, without employing any actuators. Fig.~\ref{fig2:desired_behavior} shows the desired behavior of the wrist, highlighting the desired precision (high stiffness) and compliant (high compliance) modes. When in contact with the environment, the wrist should deform to mitigate unexpected reaction forces and then return to its original shape, exhibiting self-recovery to accommodate repeated sequential impacts. Our goal is to obtain deflection without any rotation in the axial direction. To reduce added inertia and ensure compatibility with various robotic arms and grippers, we aim to design a compact and lightweight wrist without any additional sensors or actuators. To achieve these design goals, BiFlex comprises three main elements: 1) a novel buckling honeycomb structure, 2) a universal joint that restricts the deflection to two dimensions, and 3) enclosing top and bottom plates that ensure stability and facilitate customization for integration with the robot arm and gripper. From these elements, the honeycomb flexible structure introduces several design parameters that control the BiFlex behavior (bimodal stiffness modes and transition at a buckling point). These parameters can be selected based on the range of objects, tasks, and the specific robotic hand the BiFlex will interface with. In the following, we analyze the structure, while fabrication details are covered in Sec.~\ref{s:characterization}.

\begin{figure}
\vspace{3mm}
    \centering
    %\includesvg[width=\linewidth]{Figure_svg/fig_design.svg}
    \includegraphics[width=\linewidth]{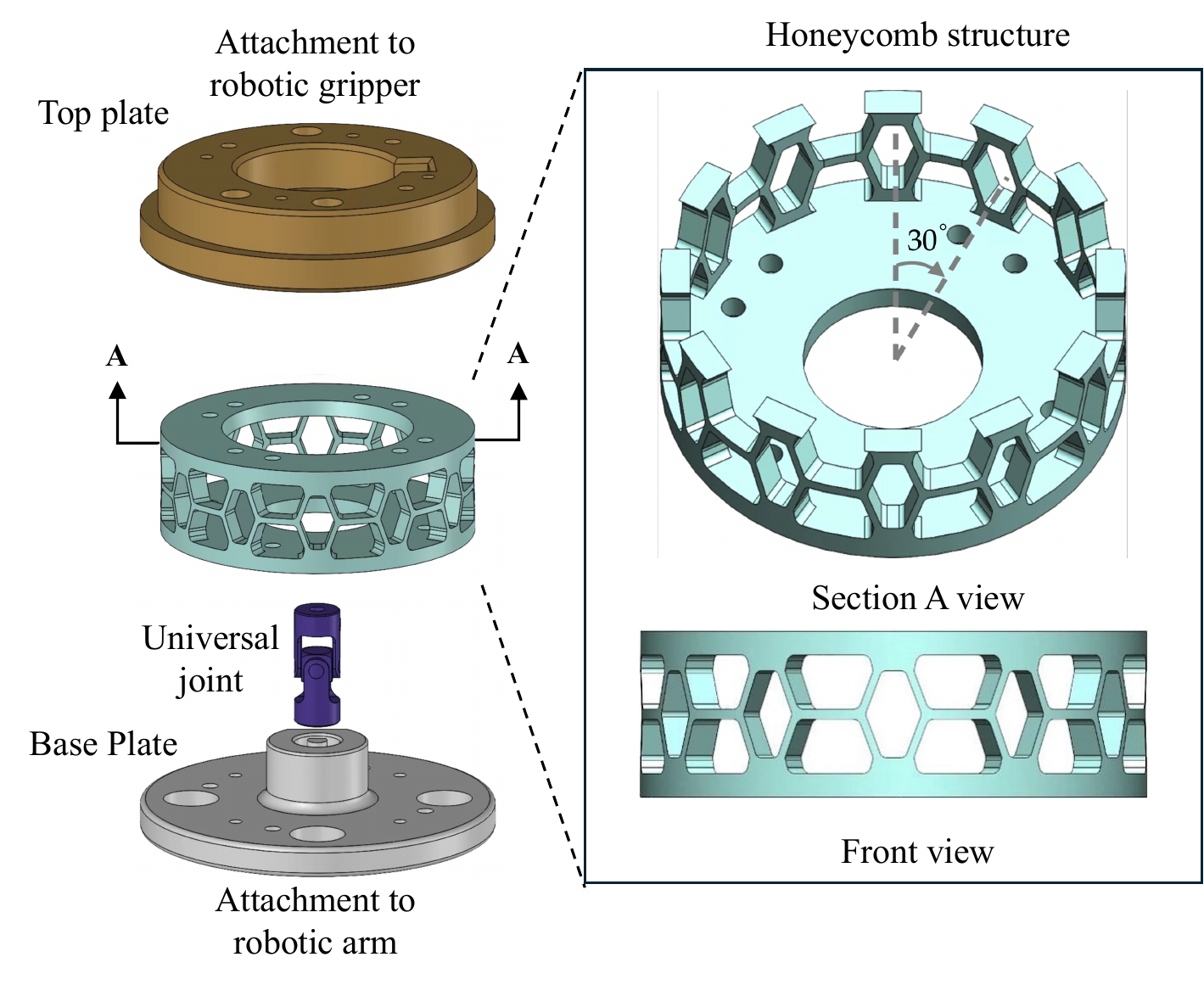}
    \caption{(left) Structural configuration of the BiFlex wrist design. The top plate and base frame are customizable to accommodate different robotic arms and grippers. (right) Cross-sectional and frontal views of the honeycomb structure.}
\vspace{-4mm}
%The U-Joint constrains unnecessary yaw motion, ensuring that load conditions and external forces are transmitted exclusively as pressing forces to the flexible wrist.  enabling the design of a flexible wrist tailored to specific gripper requirements.

\label{fig:FWD}
\end{figure}
\subsection{Buckling Honeycomb Structure}
To achieve the aforementioned design goals, we leverage the buckling effect to enable bimodal stiffness modulation. Buckling occurs when a structure under compressive load suddenly deforms, transitioning from a high-stiffness state to a lower-stiffness state at the buckling point~\cite{budiansky1974theory}. Controlling the buckling point allows us to achieve the behavior shown in Fig.~\ref{fig2:desired_behavior}, as it allows the wrist to maintain rigidity during free-space motion while becoming more compliant when interaction forces exceed a critical threshold.

To achieve uniform force distribution in all radial directions, BiFlex is comprised of twelve buckling modules evenly arranged along the edges of the base frame (Fig.~\ref{fig:FWD}B). This rotational symmetry ensures that no single module experiences disproportionate stress, leading to consistent force and torque transmission across the wrist. 
To further mitigate potential discrepancies caused by non-uniform pressing forces, we attached a universal joint   through the center of the wrist (inner diameter: 4mm; outer diameter: 9mm; height: 23mm). This universal joint eliminates unwanted yaw motion and ensures that forces are transmitted purely as compressive forces through the wrist module. To minimize the lever arm and the resulting torque from external forces, the maximum wrist height is limited to \SI{40}{mm}, including the adapter plates, with the wrist module itself occupying about \SI{21}{mm}. This compact design provides a range of motion of up to \SI{40}{\degree} and two degrees-of-freedom (flexion–extension and adduction–abduction).

Initially, we explored a single-beam wrist module, but found that this configuration could not provide the desired stiffness in the precision mode to meet both the load and deformation requirements at the buckling point. We thus adopted a honeycomb structure, inspired by the one used in corrugated cardboard. This honeycomb pattern gave us more geometric parameters to determine the bimodal stiffness behavior of the wrist~\cite{jin2024symplectic}. Specifically, the geometries of the individual honeycomb beams and the tilting angle ($\gamma$) of the honeycomb design provided an effective way to adjust the buckling point (see Fig.~\ref{fig:FWD}).

\subsection{Analysis of the Honeycomb Buckling Behavior}
\label{sec:analysis}

\begin{figure}
    \centering
\includegraphics[width=\linewidth]{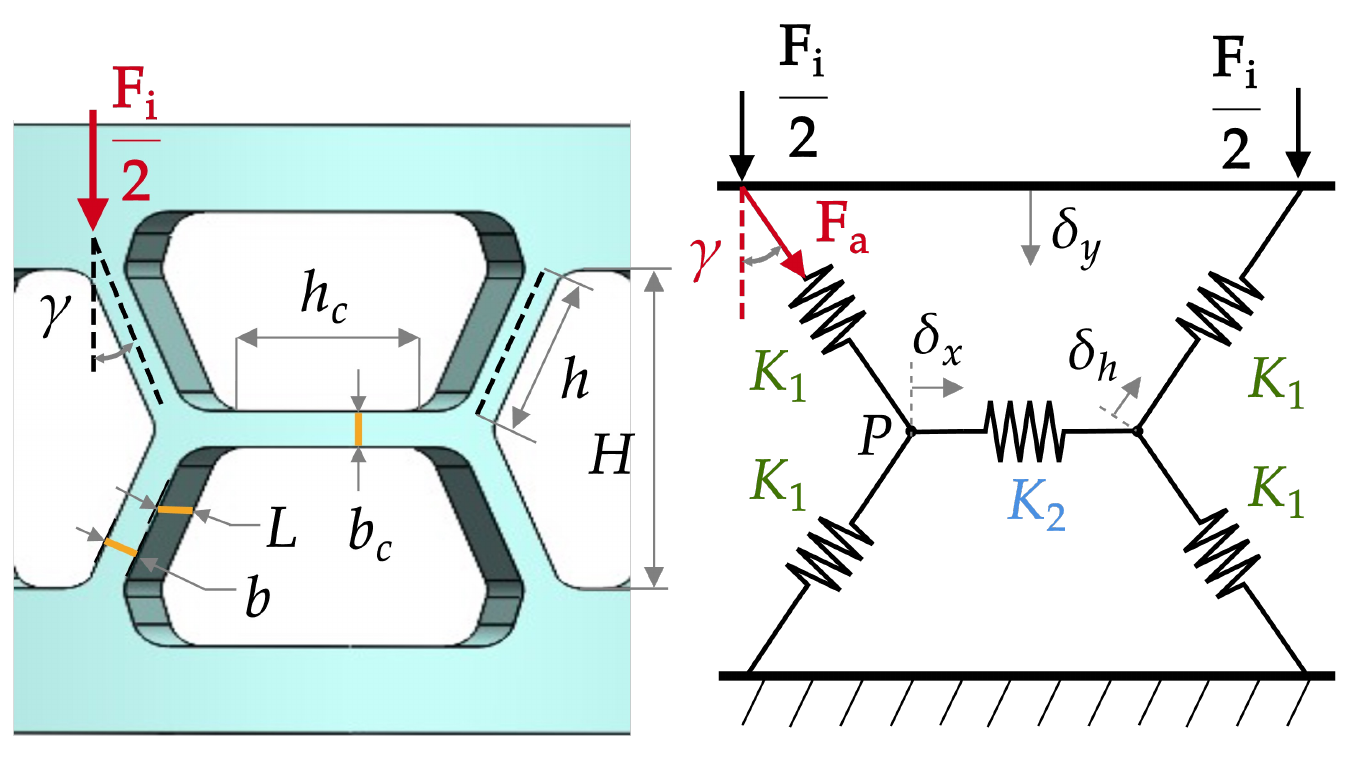}
    \caption{Frontal view (left) and free-body diagram (right) of one buckling honeycomb module. $\mathbf{F_i}$ represents the compression force applied to each single honeycomb module.}
    \label{fig:analysis}
% \vspace{-1em}
\end{figure}
To better understand when BiFlex will switch between high-stiffness precision mode and low-stiffness compliant mode, we provide the following analysis of the BiFlex honeycomb structure. We seek to understand how the geometric parameters of the honeycomb (Fig.~\ref{fig:analysis}) result in changes to the buckling point and effective stiffness of the entire honeycomb. 

Given the complexity of the 3D design, we introduced several approximations and simplifications in our analysis that may cause minor deviations from the exact behavior. In practice, we found that this analysis provided sufficient guidance on how to change the honeycomb design to create custom BiFlex wrists for a wide range of robot hands (Sec.~\ref{s:characterization} and~\ref{s:experiment}).  

%This is not intended as a fully rigorous analysis but as a guideline to how the honeycomb can be adjusted for different end effectors and desired buckling points.

% As a design guideline, not the rigorous analysis, we present a theoretical analysis of a single honeycomb module  to illustrate the trends of the key parameters influencing the bimodal stiffness and buckling response of the BiFlex without delving into detailed modeling.
We will focus on the behavior of a single honeycomb module under a pure vertical load $F_i$ (Fig.~\ref{fig:analysis}). Our first goal is to find a closed form expression for the effective vertical stiffness of the entire honeycomb module $K_{eq}$. Let $\delta_y$ be the total change in vertical displacement caused by $F_i$. By Hooke's Law, we know 
\begin{equation}
    F_i = K_{eq}\delta_y
\label{eq:hookesLaw_wholeSystem}
\end{equation}
Thus, if we can find a closed form expression for $\delta_y$, we will be able to find $K_{eq}$.

Let $\delta_h$ be the axial change in a diagonal beam and $\delta_x$ bethe horizontal displacement of point $P$ on the left side of the module. By trigonometry: $\delta_h = \delta_y\cos\gamma - \delta_x\sin\gamma$. We can find $\delta_x$ by tracing the forces from $F_i$ to the compression of the center beam. By symmetry, we will assume that the vertical load is applied equally to the left and right sections of the module, so $F_i/2$ will be transmitted into the diagonal beam as a buckling load. For simplicity, we will assume that the force component perpendicular to the diagonal beam is negligible. By Newton’s Third Law, the vertical component of $F_a$ is equal to $F_i/2$, giving us $F_a = \frac{F_i}{2\cos\gamma}$

We now turn our attention to the center beam. By symmetry, the left and right hand sides of the module transmit a horizontal force through the central beam. This will be entirely the horizontal component of $F_a$. Thus, we can express $\delta_x$ as:
\begin{equation}
    \delta_x = \frac{F_a\sin\gamma}{K_2} = \frac{F_i\sin\gamma}{2K_2\cos\gamma}
    \label{eq:deltaX}
\end{equation}

Returning to the diagonal beam, we know by Hooke's Law that $F_a = K_1\delta_h$. Substituting the definition of $\delta_h$ and our expression for $F_a$ in terms of $F_i$, we get:

\[\frac{F_i}{2\cos\gamma} = K_1 \cdot (\delta_y\cos\gamma - \delta_x\sin\gamma)\]

Solving for $\delta_y$ and substituting Eq.~\ref{eq:deltaX} for $\delta_x$, we get:
\begin{equation*}
    \delta_y = \frac{1}{\cos\gamma}\left(\frac{F_i}{2K_1\cos\gamma} + \frac{F_i\sin\gamma}{2K_2\cos\gamma}\right)
\end{equation*}

Substituting this expression for $\delta_y$ into Eq.~\ref{eq:hookesLaw_wholeSystem} and solving for $K_{eq}$:

\begin{equation}
    K_{eq} = \frac{K_2 + K_1\sin\gamma}{2K_1K_2\cos^2\gamma}
    \label{eq:K_eq}
\end{equation}

By beam equations, the axial stiffness for the diagonal and central beams, the axial stiffness is given by:

\begin{align*}
    K_1 &= \frac{EA}{h} = \frac{EbL}{h} = \frac{2EbLcos\gamma}{H}\\
    K_2 &= \frac{EA_c}{h_c} = \frac{Eb_cL}{h_c}
\end{align*}

where $E$ is the Young’s modulus of the material, $L$ is the undeformed length of the beams, $h$ is the height of the beam, and $b$ is the width of the beam, and $A$ is the cross-sectional area of the beam. No subscripts means that those are the parameters for the diagonal beams, while a subscript $c$ means that thoese are the parameters for the central beam. $H$ is the total height of the honeycomb module, so $h = \frac{H}{2\cos\gamma}$. Substituting these values into Eq.~\ref{eq:K_eq} gives a final expression of $K_eq$ in terms of the honeycomb structure's geometry:

\begin{equation}
    \boxed{K_{eq} = \frac{Hb_c + 2bh_c\cos\gamma\sin\gamma}{2ELbb_c\cos^3\gamma}}
    \label{eq:final_k_eq}
\end{equation}

Now we seek to understand when the structure will buckle. We assume that as soon as one beam buckles, the entire module will buckle together. We further assume that the side diagonal beams will buckle before the middle beam. Thus, the force required to buckle the module is the force required to buckle the diagonal beam. The equation for the critical buckling load ($P_{cr}$) for a simple beam ~\cite{lugthart_buckling_2013} is:
\begin{equation*}
P_{cr} = \frac{\pi^2 E I}{h^2} = \frac{\pi^2 E b^3 L}{12h^2}   
%= \frac{\pi^2Eb^3Lcos^2\gamma}{3H^2}
\end{equation*}
where $I$ is the second moment of area of the cross-section, which is given by $\frac{Lb^3}{12}$. 

So when $F_a = P_{cr}$, the diagonal beam will buckle. Given that $F_a = \frac{F_i}{2\cos\gamma}$, the structure will buckle when 
\begin{equation}
    \boxed{F_i \geq \frac{2\pi^2 E b^3 L\cos\gamma}{12h^2}}
    \label{eq:final_buckling}
\end{equation}

From Equations~\ref{eq:final_k_eq} and ~\ref{eq:final_buckling}, we see that the tilted angle of the diagonal beam $\gamma$ and the diagonal beam width $b$  are the primary factors affecting the location of the buckling point. This is because these factors affect the effective stiffness and force needed to buckle by a cubic factor. $\gamma$ and $b$ will be the primary ways we will change the BiFlex design for different end-effectors. %We also note that changing the middle beam stiffness (\(K_2\)) by reducing its height (\(h_c\)) provides overall system's vertical stiffness and goverens the load distribution among the components

% \section{Instantiation of the wrist design to \\ multiple robotic grippers}
% \section{Instantiation and Characterization}

\section{Instantiation to Different Robotic Grippers}
In this section, we describe how we generalize BiFlex for various robotic grippers while maintaining consistent performance. Given that over 92\% of household objects weigh less than \SI{500}{g}\cite{feix_analysis_2014}, we aim to design wrists to support objects weighing $500 \pm 50$\unit{g} with fingertip deflection under \SI{1}{cm}. As illustrated in Fig.~\ref{fig:Instron_Test_Result} (right), our design constraints are that the torque at buckling stays within a 10\% torque tolerance of the desired critical buckling load, while the angular deflection remains below the desired limit. The desired buckling point is represented by $\bigstar$, while the acceptable design range for the buckling point is represented by the hatched-line box.

We made three different versions of BiFlex for three different grippers: the Franka Hand gripper (Franka Robotics, Germany), the Robotiq 2F-85 (Robotiq, Canada), and the BaRiFlex~\cite{jeong_bariflex_2024}. The three grippers vary widely in weight and length, meaning that our desired buckling point will vary with them. To match that desired buckling point, we change $\gamma$ and $b$, as detailed in our analysis in Sec.~\ref{sec:analysis}. Table~\ref{tab:designs} compares the three grippers on their dimensions, desired buckling point and our chosen values for $\gamma$ and $b$. 

\vspace{4mm}
\setlength{\tabcolsep}{4pt} % Reduce column separation
% \vspace{-2em}
\begin{table}[]
    \centering
    \caption{Gripper characteristics and design parameters.}
    \begin{tabular}{l|cc|cc|cc}
    \toprule
    & \multicolumn{2}{c}{\textbf{Gripper dimensions}} 
    & \multicolumn{2}{|c|}{\textbf{Buckling point}} 
    & \multicolumn{2}{c}{\textbf{BiFlex parameters}} \\
    % \midrule
    & \begin{tabular}[c]{@{}c@{}} Weight \\ {[\unit{kg}]} \end{tabular} 
    & \begin{tabular}[c]{@{}c@{}} Length \\ {[\unit{mm}]} \end{tabular} 
    & \begin{tabular}[c]{@{}c@{}}Torque \\  {[\unit{Nm}]}\end{tabular} 
    & \begin{tabular}[c]{@{}c@{}}Angle\\  {[\textdegree]}\end{tabular} 
    & \begin{tabular}[c]{@{}c@{}}Width \\($b$) {[\unit{mm}]} \end{tabular} 
    & \begin{tabular}[c]{@{}c@{}}Angle  \\  ($\gamma$) {[\textdegree]}\end{tabular} 
    \\
    \midrule
    \textbf{Franka}    & 0.70 & 135 & 0.96  & 4.20  & 0.90 & 50 \\
    \textbf{Robotiq}  & 1.10 & 155 & 1.325 & 3.70  & 1.00 & 20 \\
    \textbf{BariFlex} & 0.75 & 205 & 1.378 & 2.80 & 1.20 & 5  \\
    \bottomrule
    \end{tabular}
        % \vspace{-6mm}
    \label{tab:designs}

\end{table}

\begin{figure}[t!]
    \centering
    \begin{tabular}{ccc}
    \begin{subfigure}[t]{0.3\columnwidth}
        \centering        \includegraphics[width=0.9\textwidth]{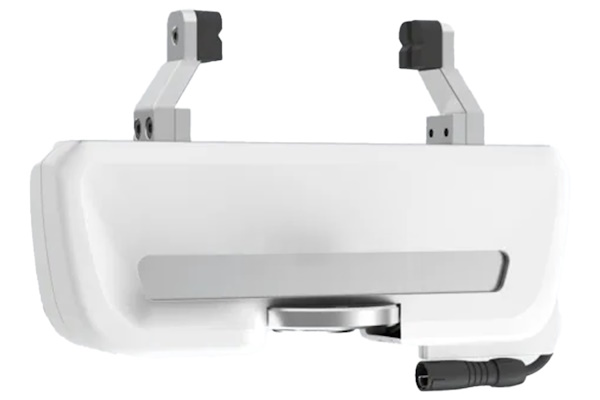}
    \end{subfigure}%
    \begin{subfigure}[t]{0.3\columnwidth}
        \centering
        \includegraphics[width=\textwidth]{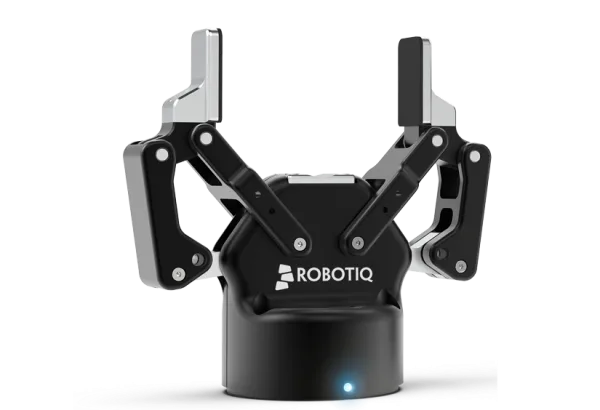}
    \end{subfigure}
        \begin{subfigure}[t]{0.3\columnwidth}
        \centering
        \includegraphics[width=\textwidth]{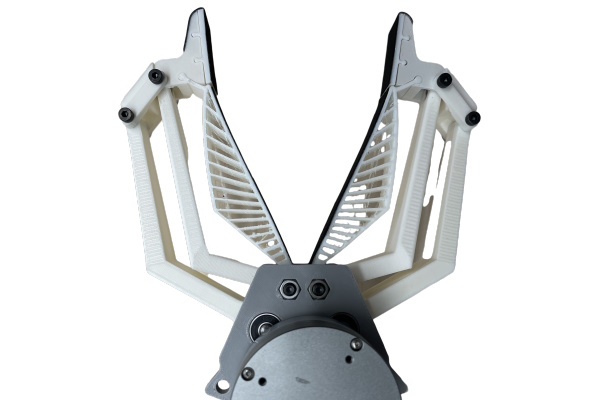}
    \end{subfigure}\\
    
    \begin{subfigure}[t]{0.3\columnwidth}
        \centering
\includegraphics[width=0.9\textwidth]{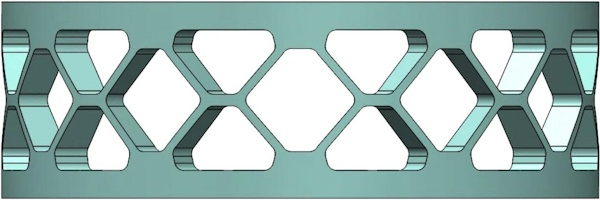}
        \caption{Franka Hand}
    \end{subfigure}%
    \begin{subfigure}[t]{0.3\columnwidth}
        \centering
        \includegraphics[width=0.9\textwidth]{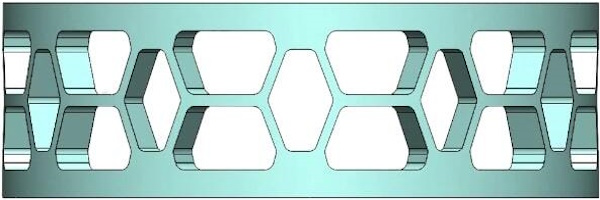}
        \caption{Robotiq 2F-85}
    \end{subfigure}
        \begin{subfigure}[t]{0.3\columnwidth}
        \centering
        \includegraphics[width=0.9\textwidth]{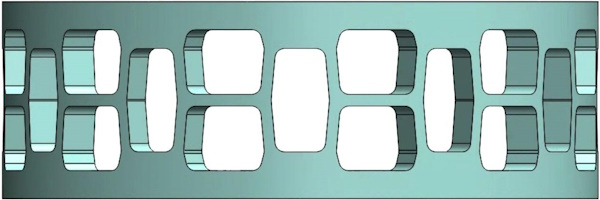}
        \caption{BariFlex \cite{jeong_bariflex_2024}}
    \end{subfigure}
    \end{tabular}
    \caption{Honeycomb structure designs corresponding to the three different grippers. Each design is tailored to the specific gripper, with the angle $\gamma$ and diagonal beam width $b$ adjusted to achieve the desired buckling point.}
    \label{fig:design_grippers}
    % \vspace{-1em}
\end{figure}

The Franka Hand gripper, being the lightest and shortest, requires the lowest buckling torque and permits the largest angle, achieved through a wide beam angle and narrow width. By contrast, the BaRiFlex gripper, being the longest, requires the highest torque and permits the smallest angle. The Robotiq gripper exhibits intermediate values, balancing beam width and angle to meet its structural requirements. The grippers and corresponding wrist designs are shown in Fig.~\ref{fig:design_grippers}.

% \subsection{Fabrication}

To achieve high stiffness with inherent compliance while minimizing weight, the wrist honeycomb modules are 3D-printed using TPU-95A (thermoplastic polyurethane), which offers a balanced combination of rigidity and flexibility at low weight. To integrate the wrist with a robot arm, we designed a base plate customized for the robotic arm and a top plate customized for the robotic gripper, both 3D-printed in PLA (polylactic acid). The entire wrist was printed in 10 hours using a Raise3D Pro3 printer, with the honeycomb module itself taking only 4 hours. This relatively short printing time underscores the simplicity and efficiency of our design, enabling rapid prototyping, testing, and maintenance.

\section{Characterization}
\label{s:characterization}
\begin{figure}[t]
\vspace{3mm}
\centering
\begin{subfigure}[c]{.30\columnwidth}
\includegraphics[trim={0.4cm 0 0.7cm 0}, clip,width=0.9\columnwidth]{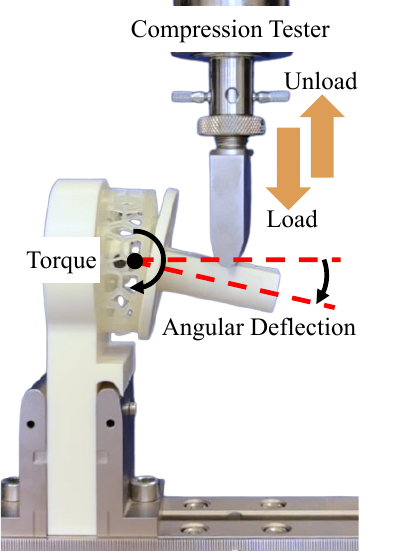}
\end{subfigure}
\begin{subfigure}[c]{.65\columnwidth}
    \includegraphics[trim={0.5cm 0 0.5cm 0},clip,width=1\columnwidth]{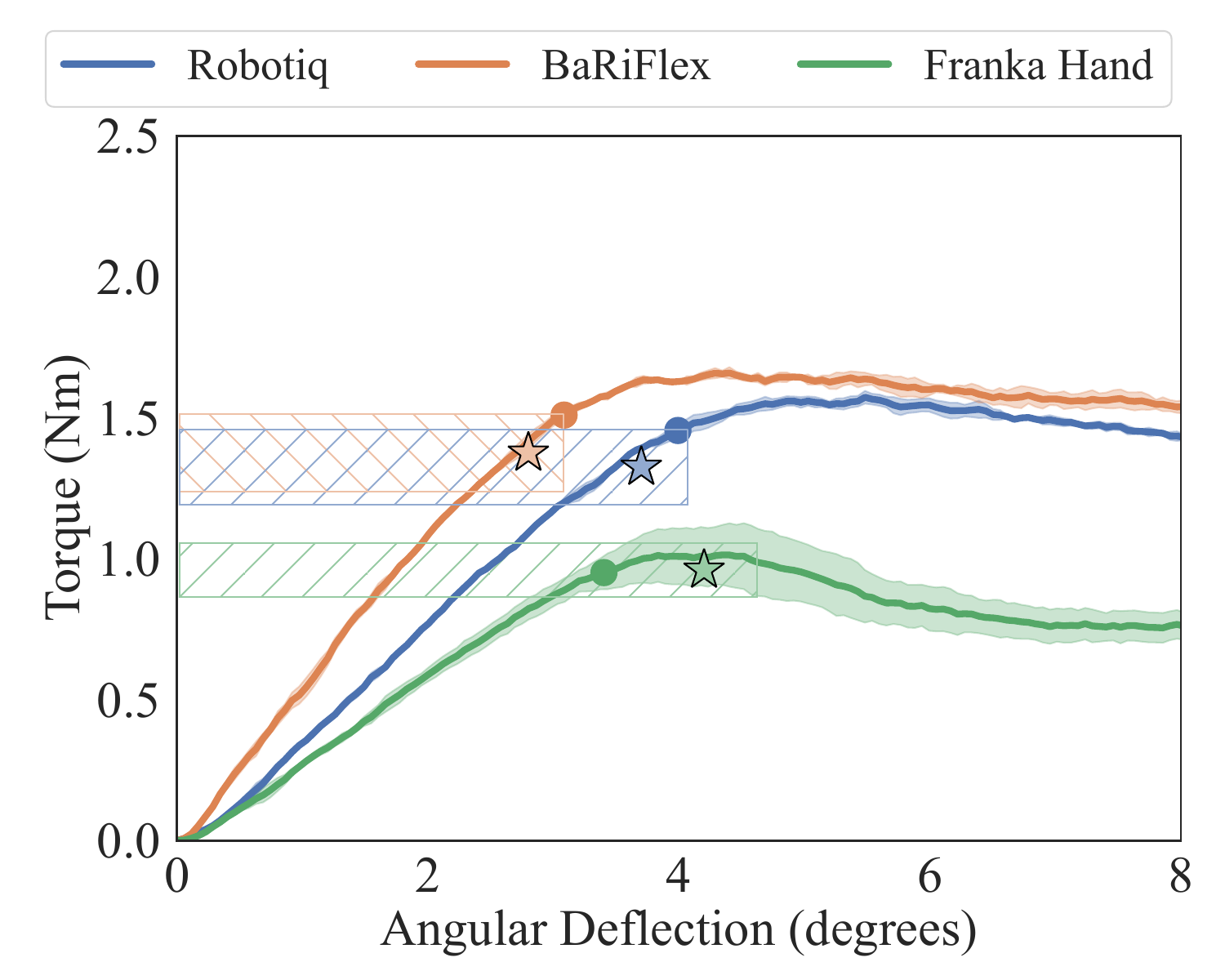}
\end{subfigure}
\caption{Compression test. (left) The compression testing machine applies a downward force on a cantilever beam connected to the BiFlex, deforming it. (right) Results indicate that the buckling point ($\bullet$) falls within its tolerance area (hatch-line box) around the desired buckling point ($\bigstar$).} % defined by a $\pm$10\% tolerance for the designed torque and a +10\% tolerance for the designed joint deflection (star). 
%Error bars represent the standard deviation over three trials.}
\vspace{-4mm}
\label{fig:Instron_Test_Result}
\end{figure}

We wish to evaluate whether the BiFlex wrists we made for each gripper successfully achieve our design criteria of two distinct stiffness modes: a high-stiffness mode for precise object manipulation and a low-stiffness mode for safe impact absorption. By comparing the measured buckling points and stiffness values against our design targets, we will confirm that our design approach can be tuned to meet the varying demands of custom grippers.

We performed a series of compression tests using a compression testing machine (Fig.~\ref{fig:Instron_Test_Result}). The machine applies a downward force on a cantilever beam, which converts the force into a wrist joint torque ($\tau$) and the pressing displacement into a wrist joint angle ($\theta$). Each wrist module was tested three times, and results were averaged.

% Figure~\ref{fig:Instron_Test_Result} illustrates the test results. The desired buckling points were determined based on each gripper’s weight and length specifications. The actual buckling point was identified as the intersection of the 80\% torque interpolation line with the horizontal line at maximum torque; this intersection, when projected onto the reference line, defines the buckling point at which the wrist transitions into its lower-stiffness mode. Notably, each measured buckling point is within 10\% of the desired value. This result indicates that decreasing the tilting angle ($\gamma$) and increasing the beam width ($b$) serve as effective means to adjust the vertical stiffness and buckling threshold for different gripper requirements. The bimodal flexible wrist for BaRiFlex shows the highest stiffness and buckling torque due to the narrow $\gamma$ angle (\SI{5}{^\circ} and thicker beam design (\SI{1.2}{mm}). In contrast, Franka Hand biomodal stiffness wrist shows the lowest stiffness.

Before the buckling point, the cantilever beam deflects and the torque rises nearly linearly, indicating a high-stiffness regime suitable for precise motion tasks. As the torque reaches its peak -- shown by the horizontal maximum torque line -— the wrist transitions into a lower-stiffness mode at the buckling point. We identify the buckling point as the intersection where the 80\% torque interpolation line meets the maximum torque plateau (projected onto the reference line). Beyond that intersection, the torque remains nearly constant even as the pressing machine continues loading, confirming that the wrist has buckled and is now in its low-stiffness mode. 

All measured values fall within the bounds of our desired buckling points described in Tab.~\ref{tab:designs}. The measured values for each BiFlex design were: \SI{3.06}{^\circ}, \SI{1.51}{Nm} for BaRiFlex; \SI{3.99}{^\circ}, \SI{1.45}{Nm} for Robotiq; and \SI{3.40}{^\circ}, \SI{0.95}{Nm} for the Franka hand, closely matching their respective targets. Furthermore, the BiFlex wrist for BaRiFlex , featuring a narrow (\SI{5}{^\circ}) and a thicker beam (\SI{1.2}{mm}), exhibits the highest stiffness and buckling torque, whereas the Franka Hand wrist demonstrates the lowest stiffness. This is in line with our predictions from Sec.~\ref{sec:analysis} of the importance of $\gamma$ and $b$. Overall, the bimodal stiffness design can be readily adapted to meet various robotic hand specifications by adjusting these key geometric parameters.

%%%%%%%%%%%%%%%%%%%%%%%
% \input{04_Characterization}
%%%%%%%%%%%%%%%%%%%%%%%
\input{Figure_tex/fig_pressing_test2}

\section{Experimental Evaluation}
\label{s:experiment}

\input{Figure_tex/fig_Wiping_test}

BiFlex has three main target features that aim to facilitate manipulation in unstructured environments: the ability to (1) enhance safety during contact interactions, (2) simplify and facilitate contact-rich manipulation tasks, and (3) maintain accuracy during object pick and place tasks. We present three functional experiments to evaluate how well BiFlex achieves these features, followed by a final demonstration of BiFlex exploiting environmental constraints to grasp flat objects.

\subsection{Mitigation of Contact Wrenches: Pressing Test}

%One of the goals of BiFlex is to increase safety and reduce possible damage to robot arms manipulating in uncertain unstructured environments.
%To that end, BiFlex absorbs some contact wrenches that appear naturally during motion and manipulation and transforms them into mechanical deformation. 
Target Feature (1) of BiFlex is to enhance safety during contact interactions. This means BiFlex should absorb the external contact wrenches that occur during normal manipulation motion to avoid potential damage to the robot arm. To evaluate BiFlex's capability of mitigating outside contact, we measure the wrenches transmitted to the robot arm by the three instantiations of the BiFlex design when the arm exerts lateral pressure with the hand's fingertips on a surface.

In this experiment, each of the BiFlex instantiations and the corresponding robotic hands are attached to a 6-axis force-torque sensor (SensONE, \textit{Bota Systems}) connected to a Franka Emika robot arm (Fig.~\ref{fig:Pressing_Test}, top).
The robot initially places the hand's fingertips \SI{5}{mm} above an aluminum frame and initiates a descending motion, applying force on the frame.
The motion continues until BiFlex deflects by \SI{10}{^\circ} and then initiates an ascending motion.
We evaluate whether the torques transmitted to the robot stay within a safety margin (\SI{15}{\newton}) and if the buckling point deformation of BiFlex maintains the transmitted wrenches under the designed values.
The measured buckling point is determined the same way as in the compression test characterizations. Each motion is repeated three times, and results are averaged.

Fig.~\ref{fig:Pressing_Test}-bottom shows the relationship between the wrist torque and deflection angle for the three robotic grippers, each paired with its corresponding BiFlex, during our experiments. The designed ($\bigstar$) and measured ($\bullet$) buckling points are indicated, along with a 10\% tolerance zone around the designed buckling point.

We observe that the measured BiFlex buckling points for the Franka Hand and Robotiq hands are within the 10\% tolerance of the designed points, indicating that the transmitted wrenches remain within safety limits. However, we observe a discrepancy between the designed and measured buckling points for the BaRiFlex hand. We hypothesize that this discrepancy may be due to the BaRiFlex being a soft robotic hand. There may be additional deformations within the BaRiFlex's design that affect the overall end-effector stiffness. A stiffer wrist design could potentially compensate for this deformation.

%We believe that this is the effect of an unmodeled deformation of the BaRiFlex itself. In fact, BaRiFlex is composed of 3D-printed PLA material and underactuated joints that introduced some additional deformation not considered during our design. 
%If we subtract the effect of the hand deformation from our results by geometrically observing the actual deformation if BaRiFlex would be completely rigid (magenta in Fig.~\ref{fig:Pressing_Test}): in this case, the measured bucking point corresponds to our design. 
% Data analysis from previous characterization tests indicates that the inherent stiffness values vary among the different wrists. However, the stiffness ranking is modified by the innate compliance of the BaRiFlex design, resulting in the order where the Robotiq wrist exhibits the highest stiffness, followed by the BaRiFlex wrist, and then the Franka Hand wrist. This is due to the innate compliance of the BaRiFlex gripper. Additionally, the test results with the gripper show a reduced buckling point value compared to the characterization tests, a reduction that is likely due to the inaccuracies of force conversion used to estimate the U-Joint torque from the fingertip force.

These results demonstrate that BiFlex effectively reduces the transmission of hazardous interaction forces from robot hands to the robot arm. By adjusting the buckling point for each hand, the BiFlex design ensures that the wrist transitions to a low stiffness mode when interaction forces exceed the specified threshold. This feature adds safety during manipulation tasks in unstructured environments, where traditional whole-rigid robot arms might fail.

% exhibits distinct buckling behavior under practical conditions with an actual gripper, compensating for contact forces by adjusting its stiffness. Furthermore, the universality of the BiFlex is confirmed by its adaptability to different grippers through appropriate design modifications.

\subsection{Adaptation in Contact-Rich Interaction: Wiping Test}

%BiFlex aims to simplify contact-rich interactions even under uncertainty.
%One of the most complex contact-rich manipulation control tasks is to wipe surfaces.
Target Feature (2) of BiFlex is to facilitate contact-rich manipulation tasks. We test whether BiFlex achieves this feature by conducting the contact-rich task of wiping a surface. In this experiment, we compare how well a robot arm wipes a tilted-surface, both with and without BiFlex. 

We use a Robotiq hand and mount it on a Franka Hand robot arm equipped with the same 6-axis force-torque sensor of the previous experiments (Fig.~\ref{fig:Wiping_Test}, left).
The gripper holds a \SI{5}{mm} sponge that will be used for wiping a triangular-shaped flat surface.
The surface presents a constant slope (\SI{30}{^\circ}), but we change its length, leading to different maximum heights.
A position controller with high stiffness moves the arm to touch the surface and slide, attempting to overcome the triangle peak.
The robot arm uses the same control strategy to wipe all triangular surfaces.
Reactive forces and torques are measured during the wiping task: if a maximum safety value is exceeded (\SI{15}{\newton}), the experiment terminates.
We compare the effect of using BiFlex vs. just a regular rigid connection.

% To evaluate the robotic wrist's ability to simplify control on uneven surfaces, we conducted a wiping experiment on a triangular hill. As illustrated in Fig.~\ref{fig:Wiping_Test}, the system is designed to validate how the BiFlex adapts to reaction forces during its interaction with a non-flat triangular surface.

% A standard height is established so that the sponge lightly contacts the peak of the triangular hill, as indicated by a small reaction force (around \SI{1.5}{N}) measured by the FT sensor. After performing the wiping test three times at this height, the robot arm is incrementally lowered by \SI{10}{mm}, thereby increasing the gap between the hill’s peak and the gripper’s height. This process continues until the measured fingertip force exceeds the threshold of \SI{15}{N}. For comparison, the experiment is also conducted using a rigid wrist (i.e., without the BiFlex) to assess the maximum wiping height and further validate the proposed wrist’s adaptation performance on non-flat surfaces under position control.

Fig.~\ref{fig:Wiping_Test} depicts the reaction forces (middle row) and torques (bottom row) as a function of horizontal end-effector displacement on the wiping surface for both the BiFlex (middle column) and the rigid wrist (right column) during the wiping experiment. 
The top row in Fig.~\ref{fig:Wiping_Test} displays the shape of the triangular surface in each experiment. 
We observe that the robot equipped with BiFlex can successfully wipe triangular shapes of up to \SI{50}{mm} height while keeping the reaction force below the \SI{15}{N} safety threshold.
Buckling only occurs for triangles of \SI{20}{\milli\meter} or higher, using the high stiffness regime for lower triangular shapes.
% When attempting to wipe a \SI{55}{mm} hill, the reaction force reaches the threshold as the gripper passes the summit. 
% No buckling behavior is observed for hill heights between \SI{0}{mm} and \SI{15}{mm}, since the applied force remains below the buckling limit; however, for surfaces higher than \SI{15}{mm}, buckling behavior occurs. 
% The green dot in the middle left figure marks the buckling point, which is observed when the wrist experiences a reaction force of approximately \SI{11}{N}. Beyond this point, the reaction force decreases even as the surface height increases and the gripper approaches the peak of the hill. In the graphs for the \SI{50}{mm} and \SI{55}{mm} conditions, the reaction force increases again, with the \SI{55}{mm} case reaching the threshold as the BiFlex becomes fully compressed.
By contrast, the results obtained with the rigid wrist reveal that, without any compliant wrist element, the reaction force and torque increase dramatically as the height of the triangular shape increases. 
The rigid wrist is able to wipe a surface with a hill height of up to \SI{14}{mm} thanks to the small deformation enabled by the sponge, but the reaction force soon exceeds the \SI{15}{N} safety threshold when the hill height surpasses \SI{15}{mm}.

Our results indicate that the BiFlex enables safe and simple-to-control contact-rich interactions, even under changing task conditions (height of the surface to wipe). 
We believe this is a significant advantage for manipulation in unstructured environments over complex force control strategies that may require active force feedback: BiFlex enables safe and successful wiping execution with a simple high-stiff position controller.
% Its inherent flexibility not only simplifies control but also enhances performance across a wide range of surface conditions.
%Furthermore, the flexibility inherent in its design enhances its performance under varying surface conditions.

\subsection{Maintaining Accuracy: Pick-and-Place Test}
\input{Figure_tex/fig_pick_and_place}
Target Feature (3) of BiFlex is to maintain positional accuracy during pick and place tasks. We see if we achieve this feature by conducting a pick and place task with 15 different household objects of varying weights (Fig.~\ref{fig:Pick_and_place_test}-Left). We designed BiFlex to support objects of up to \SI{500}{g} at the fingertip before it buckles, so we should maintain precision for objects below that weight limit. 

In the experiment, a Robotiq gripper was mounted on the BiFlex, which was attached to a robotic arm. The test environment consisted of a three-tier cabinet with the target object initially placed on the second shelf. The robotic arm approached the object to grasp it, then lifted it to a designated position where the object was held \SI{10}{mm} above the third shelf without inducing wrist deflection, and finally moved slowly into the cabinet to place the object. The height of the third shelf was adjusted by $\pm$\SI{10}{mm} relative to the object’s height to introduce controlled variation. Each object underwent three pick-and-place attempts, and any instance in which the object contacted the cabinet shelf due to deformation from the buckling effect was recorded as a failure.

14 objects weighing less than \SI{500}{g} were successfully picked up and placed at their designated locations, as shown in Fig.~\ref{fig:Pick_and_place_test} below. However, when the gripper attempted to lift a \SI{600}{g} water bottle, the BiFlex was unable to withstand the resulting torque, causing it to buckle. In the post-buckling state, the fingertip deflected by more than \SI{10}{mm} and the force-torque sensor recorded high forces as the object contacted the cabinet floor. These results demonstrate that our bimodal stiffness design—leveraging a controlled buckling behavior—is both precise and robust, enabling accurate pick-and-place operations with various household objects.

\subsection{Grasping with Environmental Constraints}
\input{Figure_tex/fig_sliding_demonstration}
As discussed in Sec.~\ref{s:intro}, manipulating in unstructured environments, inherently uncertain, can be significantly simplified by exploiting environmental constraints. Without them, grasping a thin object on a table would require highly precise vertical positioning: small perception or positioning errors would cause grasp failures or the gripper to collide with the surface and generate large reaction forces. 
BiFlex enables direct and safe use of environmental constraints for grasping and manipulation, as we demonstrate in the last experiment: we command the robot to execute a grasping motion on a small object (a hex wrench) using the table as constraint, i.e., sliding on it with the fingers before grasping. 
To simulate possible perceptual inaccuracies, we command the robot to move to different heights: from the exact height of the object to reaching \SI{50}{mm} below the table surface in increments of \SI{5}{mm}, which allows it to use the table as a constraint.
The robot with BiFlex succeeds in all cases.
In comparison, without BiFlex the robot can only grasp when commanded to move to the exact height of the object and \SI{5}{mm} below; for all other heights, the forces exceed the safety limits.
Fig.~\ref{fig_sliding_demonstration} depicts the execution when the hand is commanded to move \SI{50}{mm} below the table, and the reaction forces and torques, which are kept under safety margins thanks to BiFlex.
After the wrench is lifted, the BiFlex reverts to its original shape. 
This test indicates that, when equipped with BiFlex, the robot can overcome inaccuracies in grasping height, leveraging environmental constraints for robust manipulation and safely mitigating excessive contact forces.

% Figure~\ref{fig_sliding_demonstration} illustrates a scenario where a Robotiq arm, tilted \SI{30}{^\circ} relative to the table, attempts to grasp a hex wrench. The arm is commanded to a desired position \SI{50}{mm} below the object’s actual height. As the fingertip of the gripper touches the surface and continues moving downward, the X-directional reaction force (blue line, $F_x$) remains below \SI{11}{N} thanks to the buckling effect of the BiFlex. After the wrench is lifted, the BiFlex reverts to its original shape. This demonstration confirms that BiFlex can accommodate height inaccuracies by effectively mitigating excessive contact forces, enabling safer and more robust grasping even under positional errors.

% \begin{figure}[htbp]
% \centering
% \begin{subfigure}[b]{0.47\columnwidth}
%    \includegraphics[width=\columnwidth]{Figure/Drawing_PickAndPlace_test1.jpg}
%    \caption{}
%    \label{fig:PickAndPlaceTest1}
% \end{subfigure}
% % \hfill
% \begin{subfigure}[b]{0.48\columnwidth}
%    \includegraphics[width=1.0\columnwidth]{Figure/Object_PickAndPlace.JPG}
%    \caption{}
%    \label{fig:PickAndPlaceTest2}
% \end{subfigure}
% \caption{Pick and Place}
% \label{fig:PickAndPlaceTest}
% \end{figure}

%%%%%%%%%%%%%%%%%%%%%%%
% \input{06_Result}
%%%%%%%%%%%%%%%%%%%%%%%
\section{Discussion and Conclusion}
\label{s:discussion}

% \begin{itemize}
%     \item Advantages Over Existing Designs: Discuss how our design outperforms others in terms of simplicity, cost-effectiveness, and performance in contact-rich environments.
%     \item Limitations: Address the observed limitations, such as low-frequency oscillations during the wiping test due to stress relaxation, and analyze their impact on overall performance.
%     \item Impact on Applications and Practicality: Evaluate the broader implications of our design on real-world applications, including its potential for improved adaptability, robustness, and practical implementation in dynamic environments.
% \end{itemize}

% Close again with a general discussion in the lines of embodiment computation, how this kind of alternative embodiments simplify robot manipulation in unstructured environments making it easier and safer.

The BiFlex wrist design offers significant advantages over both fully actuated and traditional compliant wrists. Its passive mechanism—built around a 3D-printed honeycomb structure—eliminates the need for extra actuators, sensors, and complex control algorithms. This simplicity not only reduces cost but also facilitates seamless integration with a variety of robotic arms and grippers. By leveraging the buckling effect, the design achieves reliable bimodal stiffness: high precision during free-space manipulation and safe compliance during contact-rich interactions, thereby simplifying control and improving performance under unexpected loads. 

Despite these strengths, the BiFlex wrist does exhibit some limitations. First, we observe low-frequency oscillations during the contact-rich (wiping) tests, likely caused by stress relaxation in the compliant materials, which could cause minor deviations from the desired force profile and affect consistent contact performance. Further refinement of the structure or alternative material choices would be necessary to mitigate them.
We also observe some deviation from the designed parameters due to tolerances in the U-Joint component.
Finally, as any robotic soft component, tear and wear can change the properties of BiFlex over time. However, given the simplicity of its manufacturing, this can be alleviated by replacing the honeycomb structure.
% Overall, BiFlex's passive, bimodal stiffness capability allows the wrist to naturally adjust its compliance based on external loads, making it particularly valuable for object manipulation, surface wiping, and safe human–robot interactions. Its universal compatibility and reduced complexity pave the way for broader application in unstructured, contact-rich environments.

In summary, the BiFlex wrist employs an embodied, passive design to achieve bimodal stiffness through a 3D-printed honeycomb structure and buckling mechanism—simplifying control, lowering costs, and enhancing safety and adaptability in dynamic, unpredictable settings.
Its novel bimodal stiffness capability allows the wrist to naturally adjust its compliance based on external loads, making it particularly valuable for object manipulation, surface wiping, and potentially safe human–robot interactions in unstructured environments.
%%%%%%%%%%%%%%%%%%%%%%%
% \input{08_Conclusion}

\bibliography{REF2}

\begin{thebibliography}{10}

\bibitem{della2017postural}
C.~Della~Santina, M.~Bianchi, G.~Averta, S.~Ciotti, V.~Arapi, S.~Fani, E.~Battaglia, M.~G. Catalano, M.~Santello, and A.~Bicchi, ``Postural hand synergies during environmental constraint exploitation,'' {\em Frontiers in neurorobotics}, vol.~11, p.~41, 2017.

\bibitem{mason1985mechanics}
M.~Mason, ``The mechanics of manipulation,'' in {\em Proceedings. 1985 IEEE International Conference on Robotics and Automation}, vol.~2, pp.~544--548, IEEE, 1985.

\bibitem{erdmann1988exploration}
M.~A. Erdmann and M.~T. Mason, ``An exploration of sensorless manipulation,'' {\em IEEE Journal on Robotics and Automation}, vol.~4, no.~4, pp.~369--379, 1988.

\bibitem{eppner2015exploitation}
C.~Eppner, R.~Deimel, J.~Alvarez-Ruiz, M.~Maertens, and O.~Brock, ``Exploitation of environmental constraints in human and robotic grasping,'' {\em The International Journal of Robotics Research}, vol.~34, no.~7, pp.~1021--1038, 2015.

\bibitem{kanitz_compliant_2018}
G.~Kanitz, F.~Montagnani, M.~Controzzi, and C.~Cipriani, ``Compliant {Prosthetic} {Wrists} {Entail} {More} {Natural} {Use} {Than} {Stiff} {Wrists} {During} {Reaching}, {Not} ({Necessarily}) {During} {Manipulation},'' {\em IEEE Transactions on Neural Systems and Rehabilitation Engineering}, vol.~26, pp.~1407--1413, July 2018.

\bibitem{siciliano2008springer}
B.~Siciliano, O.~Khatib, and T.~Kr{\"o}ger, {\em Springer handbook of robotics}, vol.~200.
\newblock Springer, 2008.

\bibitem{siciliano2009force}
B.~Siciliano, L.~Sciavicco, L.~Villani, and G.~Oriolo, {\em Force control}.
\newblock Springer, 2009.

\bibitem{khatib1987unified}
O.~Khatib, ``A unified approach for motion and force control of robot manipulators: The operational space formulation,'' {\em IEEE Journal on Robotics and Automation}, vol.~3, no.~1, pp.~43--53, 1987.

\bibitem{hogan1984impedance}
N.~Hogan, ``Impedance control: An approach to manipulation,'' in {\em 1984 American control conference}, pp.~304--313, IEEE, 1984.

\bibitem{montagnani2015finger}
F.~Montagnani, M.~Controzzi, and C.~Cipriani, ``Is it finger or wrist dexterity that is missing in current hand prostheses?,'' {\em IEEE Transactions on Neural Systems and Rehabilitation Engineering}, vol.~23, no.~4, pp.~600--609, 2015.

\bibitem{phan2020estimating}
G.-H. Phan, C.~Hansen, P.~Tommasino, A.~Budhota, D.~M. Mohan, A.~Hussain, E.~Burdet, and D.~Campolo, ``Estimating human wrist stiffness during a tooling task,'' {\em Sensors}, vol.~20, no.~11, p.~3260, 2020.

\bibitem{bajaj_state_2019}
N.~M. Bajaj, A.~J. Spiers, and A.~M. Dollar, ``State of the {Art} in {Artificial} {Wrists}: {A} {Review} of {Prosthetic} and {Robotic} {Wrist} {Design},'' {\em IEEE Transactions on Robotics}, vol.~35, pp.~261--277, Feb. 2019.

\bibitem{fan_prosthetic_2022}
H.~Fan, G.~Wei, and L.~Ren, ``Prosthetic and robotic wrists comparing with the intelligently evolved human wrist: {A} review,'' {\em Robotica}, vol.~40, pp.~4169--4191, Nov. 2022.
\newblock Publisher: Cambridge University Press.

\bibitem{zhang_stiffness_2020}
Z.~Zhang, G.~Chen, W.~Fan, W.~Yan, L.~Kong, and H.~Wang, ``A {Stiffness} {Variable} {Passive} {Compliance} {Device} with {Reconfigurable} {Elastic} {Inner} {Skeleton} and {Origami} {Shell},'' {\em Chinese Journal of Mechanical Engineering}, vol.~33, p.~75, Dec. 2020.

\bibitem{von2020compact}
F.~von Drigalski, K.~Tanaka, M.~Hamaya, R.~Lee, C.~Nakashima, Y.~Shibata, and Y.~Ijiri, ``A compact, cable-driven, activatable soft wrist with six degrees of freedom for assembly tasks,'' in {\em 2020 IEEE/RSJ International Conference on Intelligent Robots and Systems (IROS)}, pp.~8752--8757, IEEE, 2020.

\bibitem{zhang_0_2024}
Q.~Zhang, Z.~Hu, W.~Wan, and K.~Harada, ``{Compliant} {Peg}-in-{Hole} {Assembly} {Using} a {Very} {Soft} {Wrist},'' {\em IEEE Robotics and Automation Letters}, vol.~9, pp.~17--24, Jan. 2024.

\bibitem{formica2012passive}
D.~Formica, S.~K. Charles, L.~Zollo, E.~Guglielmelli, N.~Hogan, and H.~I. Krebs, ``The passive stiffness of the wrist and forearm,'' {\em Journal of neurophysiology}, vol.~108, no.~4, pp.~1158--1166, 2012.

\bibitem{borzelli2018muscle}
D.~Borzelli, B.~Cesqui, D.~J. Berger, E.~Burdet, and A.~d’Avella, ``Muscle patterns underlying voluntary modulation of co-contraction,'' {\em PLoS One}, vol.~13, no.~10, p.~e0205911, 2018.

\bibitem{sun_compact_2025}
H.~Sun, S.~Park, and D.~Hwang, ``Compact {Modular} {Robotic} {Wrist} {With} {Variable} {Stiffness} {Capability},'' {\em IEEE Transactions on Robotics}, vol.~41, pp.~141--158, 2025.

\bibitem{milazzo2024modeling}
G.~Milazzo, M.~G. Catalano, A.~Bicchi, and G.~Grioli, ``Modeling and control of a novel variable stiffness three dofs wrist,'' {\em The International Journal of Robotics Research}, vol.~43, no.~12, pp.~1898--1915, 2024.

\bibitem{montagnani_preliminary_2013}
F.~Montagnani, M.~Controzzi, and C.~Cipriani, ``Preliminary design and development of a two degrees of freedom passive compliant prosthetic wrist with switchable stiffness,'' in {\em 2013 {IEEE} {International} {Conference} on {Robotics} and {Biomimetics} ({ROBIO})}, pp.~310--315, Dec. 2013.

\bibitem{budiansky1974theory}
B.~Budiansky, ``Theory of buckling and post-buckling behavior of elastic structures,'' {\em Advances in applied mechanics}, vol.~14, pp.~1--65, 1974.

\bibitem{jin2024symplectic}
M.~Jin, X.~Hou, W.~Zhao, and Z.~Deng, ``Symplectic stiffness method for the buckling analysis of hierarchical and chiral cellular honeycomb structures,'' {\em European Journal of Mechanics-A/Solids}, vol.~103, p.~105164, 2024.

\bibitem{lugthart_buckling_2013}
J.~Lugthart, ``Buckling of a {Beam},'' {\em Leiden University}, Nov. 2013.

\bibitem{feix_analysis_2014}
T.~Feix, I.~M. Bullock, and A.~M. Dollar, ``Analysis of {Human} {Grasping} {Behavior}: {Object} {Characteristics} and {Grasp} {Type},'' {\em IEEE Transactions on Haptics}, vol.~7, pp.~311--323, July 2014.

\bibitem{jeong_bariflex_2024}
G.-C. Jeong, A.~Bahety, G.~Pedraza, A.~D. Deshpande, and R.~Martín-Martín, ``{BaRiFlex}: {A} {Robotic} {Gripper} with {Versatility} and {Collision} {Robustness} for {Robot} {Learning},'' in {\em 2024 {IEEE}/{RSJ} {International} {Conference} on {Intelligent} {Robots} and {Systems} ({IROS})}, (Abu Dhabi, United Arab Emirates), pp.~4106--4113, IEEE, Oct. 2024.

\end{thebibliography}
\bibliographystyle{ieeetr}

%%%%%%%%%%%%%%%%%%%%%%%%%%%%%%%%%%%%%%%%%%%%%%%%%%%%%%%%%%%%%%%%%%%%%%%%%%%%%%%%

\end{document}